\documentclass[10pt,letterpaper]{article}
\usepackage[top=0.85in,left=2.75in,footskip=0.75in,marginparwidth=2in]{geometry}

\usepackage[utf8]{inputenc}

\usepackage{cite}

\usepackage{algorithm2e}

\usepackage{nameref,hyperref}

\usepackage[right]{lineno}

\usepackage{microtype}
\DisableLigatures[f]{encoding = *, family = * }

\raggedright
\usepackage{changepage}

\usepackage[aboveskip=1pt,labelfont=bf,labelsep=period,singlelinecheck=off]{caption}

\makeatletter
\renewcommand{\@biblabel}[1]{\quad#1.}
\makeatother

\usepackage{lastpage,fancyhdr,graphicx}
\usepackage{epstopdf}
\fancyheadoffset[L]{2.25in}
\fancyfootoffset[L]{2.25in}

\usepackage{color}

\definecolor{Gray}{gray}{.25}

\usepackage{graphicx}
\usepackage{multirow}
\usepackage{amsmath}
\usepackage{amssymb}

\usepackage{sidecap}

\usepackage{wrapfig}
\usepackage[pscoord]{eso-pic}
\usepackage[fulladjust]{marginnote}
\reversemarginpar

\begin{document}
\vspace*{0.35in}


\begin{flushleft}
{\Large
\textbf\newline{Metaheuristic Algorithms for Convolution Neural Networks}
}
\newline
\\
L.M. Rasdi Rere\textsuperscript{1,2},
Mohamad Ivan Fanany\textsuperscript{1},
Aniati Murni Arymurthy\textsuperscript{1},
\\
\bigskip
\bf{1} Machine Learning and Computer Vision Laboratory, \\Faculty of Computer Science, Universitas Indonesia\\
\bf{2} Computer System Laboratory, STMIK Jakarta STI\&K\\
\bigskip
* laode.mohammad@ui.ac.id

\end{flushleft}

\providecommand{\keywords}[1]{\textbf{\textit{Keywords---}} #1}

\section*{Abstract}
A typical modern optimization technique is usually either heuristic or metaheuristic. This technique has managed to solve some optimization problems in the research area of science, engineering, and industry. However, implementation strategy of metaheuristic for accuracy improvement on convolution neural networks (CNN), a famous deep learning method, is still rarely investigated.  Deep learning relates to a type of machine learning technique, where its aim is to move closer to the goal of artificial intelligence of creating a machine that could successfully perform any intellectual tasks that can be carried out by a human. In this paper, we propose the implementation strategy of three popular metaheuristic approaches, i.e. simulated annealing, differential evolution and harmony search, to optimize CNN. The performance of these metaheuristic methods in optimizing CNN on classifying MNIST and CIFAR dataset were evaluated and compared. Furthermore, the proposed methods are also compared with the original CNN. Although the proposed methods show an increase in the computation time, their accuracy has also been improved (up to 5.73 percent).
\bigskip

\noindent\keywords{metaheuristic, convolution neural network, deep learning, simulated annealing, differential evolution, harmony search}


\section*{Introduction}
\label{Introduction}

Deep learning (DL) is mainly motivated by the research of artificial intelligent, in which the general goal is to imitate the ability of human brain to observe, analyze, learn and make a decision, especially for complex problem \cite{Naja}. This technique is in the intersection amongst the research area of signal processing, neural network, graphical modeling, optimization and pattern recognition. The current reputation of DL is implicitly due to drastically improve the abilities of chip processing, significantly decrease the cost of computing hardware and advance research in machine learning and signal processing \cite{Deng}.

In general, the model of DL technique can be classified into discriminative models, generative models, and hybrid model \cite{Deng}. Discriminative models, for instance, are CNN, deep neural networks, and recurrent neural network. Some examples of generative models are deep belief networks (DBN), restricted Boltzmann machine, regularized autoencoders, and deep Boltzmann machines. On the other hand, hybrid model refers to the deep architecture use the combination of a discriminative and generative model. An example of this model is DBN to pre-train deep CNN, which can improve the performance of deep CNN over random initialization. Among all of hybrid DL techniques, this paper focuses on metaheuristic optimization for training a CNN. 

Although the sound character of DL to solve a variety of learning tasks, training is difficult\cite{Lamos} \cite{Glauner} \cite{Rasdi}. Some examples of successful methods for training DL are Stochastic Gradient Descent, Conjugate gradient, Hessian-free Optimization and Krylov Subspace Descent.

Stochastic Gradient Descent is easy to implement and also fast in the process for a case with many training samples. However, this method needs several manual tuning to make its parameters optimal, and also its process is principally sequential, as a result, it is hard to parallelize them with GPUs. Conjugate Gradient (CG) on the other side is easier to check for convergence as well as more stable to train. Nevertheless, CG is slow, so that it needs multicore CPUs and availability of a vast number of RAMs \cite{QVLee}. 

Hessian-free optimization (HFO) has been applied to train deep auto-encoders\cite{Martens}, proficient in handling under fitting problem, and more efficient than pre-training + fine tuning proposed by Hinton and Salakhutdinov \cite{Hinton}. On the other side, Krylov Subspace Descent (KSD) is more robust and simpler than HFO as well as look like to work better for the classification performance and optimization speed. However, KSD needs more memory than HFO \cite{Vinyal}.

In fact, techniques of modern optimization are heuristic or metaheuristic. These optimization techniques have been applied to solve any optimization problems in the research area of science, engineering, and even industry \cite{Yang}. However, research about metaheuristic for optimize deep learning method is rarely conducted. One of paper is the combining of genetic algorithm (GA) and CNN, proposed by You Zhining and Pu Yunming \cite{You}.Their model select the CNN characteristic by the process of recombination and mutation on GA, in which the model of CNN exists as individual in the algorithm of GA. Besides, in recombination process, only the layers weights and threshold value of C1 (convolution in first layer)  and C3 (convolution in third layer) are changed in CNN model.

In this paper, we compared the performance of three metaheuristic algorithms, i.e. simulated annealing (SA), differential evolution (DE) and harmony search (HS), for optimizing CNN.. The strategies by looking for the best value of the fitness  function on the last layer using metaheuristic algorithm, then the results will be used again to calculate the weights and biases in the previous layer. In case of testing the performance of the proposed methods, we use MNIST dataset. This dataset is images of digital handwritten digits, in which it contains 60,000 training data and 10,000 testing data. All of the images have been centered and standardized with the size of 28 x 28 pixels. Each pixel of the image is represented by 0 for black, 255 for white and in between is a different shade of gray \cite{LISA}. 

This paper is organized as follow:  Section 1 is an introduction, Section 2 explains about the used metaheuristic algorithms, Section 3 describe the convolution neural networks, Section 4 gives a description of the proposed methods, Section 5 present result of simulation, and Section 6 is the conclusion.

\section{Metaheuristic algorithms}
Metaheuristic is well-known as an efficient method for hard optimization problems, i.e.  the problems that cannot be solved optimally using deterministic approach within a reasonable time limit. Metaheuristic methods work for three main purposes: for fast solving problem, for solving large problems, for making a more robust algorithm. These methods are also simple to design as well as flexible and easy to implement \cite{Talbi}. 

In general, metaheuristic algorithms use the combination of rules and randomization to duplicate the phenomena of nature. The biological system imitation of metaheuristic algorithm, for instance, are evolution strategy, GA, and DE. Phenomena of ethology for examples are particle swarm optimization (PSO), bee colony optimization (BCO), bacterial foraging optimization algorithms (BFOA), and ant colony optimization (ACO). Phenomena of physic are SA, microcanonical annealing and threshold accepting method \cite{Boussaid}. Another form of metaheuristic is inspired by music phenomena, such as HS algorithm \cite{Lee}. 

Classification of metaheuristic algorithm can also be divided into single-solution based and population-based. Some of the examples for single-solution based metaheuristic are the noising method, tabu search, SA, TA, and guided local search. In the case of metaheuristic based on population, it can be classified into swarm intelligent and evolutionary computation. The general term of swarm intelligent is inspired by the collective behavior of social insect colonies or animal societies. Examples of these algorithms are GP, GA, ES, and DE. On the other side, the algorithm for evolutionary computation takes inspiration from the principles of Darwinian for developing adaptation into their environment. Some examples of these algorithms are PSO, BCO, ACO, and BFOA \cite{Boussaid}. Among of all these metaheuristic algorithms, SA, DE and HS are used in this paper. 

\subsection{Simulated Annealing algorithm}
SA is a technique of random search for the problem of global optimization. It mimics the process of annealing in material processing\cite{Yang}. This technique was firstly proposed in 1983 by Kirkpatrick, Gelatt, and Vecchi \cite{Kirkpatrick}.

The principle idea of SA is using random search, which not only allows changes that improve the fitness function but also maintaining some changes that are not ideal. As example, in minimum optimization problem, any better changes that decrease the fitness function value $f(x)$ will be accepted, but some changes that increase $f(x)$ will also be accepted with a transition probability ($p$) as follow:
\begin{equation}
    p=\exp({\frac{-\Delta E}{kT}})
\end{equation}
where \( \Delta E\) is the energy level changes, $k$ is the Boltzmann's constant, and $T$ is temperature for controlling the process of annealing. This equation is based on the Boltzmann distribution in physics \cite{Yang}. The following is standard procedure of SA for optimization problems: 
\begin{enumerate}
    \item \textbf{Generate the solution vector:} The initial solution vector is randomly selected, and then the fitness function is calculated.
    \item \textbf{Initialize the temperature:} If the temperature value is too high, it will take a long time to reach convergence, whereas too small value can cause the system missed the global optimum.
    \item \textbf{Select a new solution:} A new solution is randomly selected from the neighborhood of the current solution.
    \item \textbf{Evaluate a new solution:} A new solution is accepted as a new current solution depending on its fitness function.
    \item \textbf{Decrease the temperature:} During the search process, the temperature is periodically decreased.
    \item \textbf{Stop or repeat:} The computation is stopped when the termination criterion is satisfied. Otherwise, step 2 and 6 are repeated.
\end{enumerate}

\subsection{Differential Evolution algorithm}
Differential Evolution is firstly proposed by Price and Storn in 1995, to solve the Chebyshev polynomial problem \cite{Boussaid}. This algorithm is created on individual’s difference, exploiting random search in the space of solution, and finally operate the procedure of mutation, crossover, as well as selection to obtain the suitable individual in system \cite{Noman}. 

There are some types in DE, including the classical form is DE/rand/1/bin, it indicates that in the process of mutation, the target vector is randomly selected, and only a single different vector is applied. The acronym of bin shows that crossover process is organized by a rule of binomial decision. The procedure of DE algorithm is shown by the following steps:

\begin{enumerate}
    \item \textbf{Determining parameter setting:} Population size is the number of individuals. Mutation factor (F) control the magnification of the two individual differences to avoid search stagnation. Crossover rate (CR) decides how many consecutive genes of the mutated vector are copied to the offspring.
    \item \textbf{Initialization of population:} The population is produced by randomly generating the vectors in the suitable search range. 
    \item \textbf{Evaluation of individual:} Each of individual is evaluated by calculating their objective function. 
    \item \textbf{Mutation operation:} Mutation adds identical variable to one or more vector parameters. In this operation, three auxiliary parents \((x_M^{p1}, x_M^{p2}, x_M^{p3})\) are selected randomly, in which they will participate in mutation operation to create a mutated individual \( x_M^{mut}\) as follows:
    \begin{equation}
        x_M^{mut}=x_M^{p1}+F(x_M^{p2}-x_M^{p3})
    \end{equation} 
    
    \item[] where \(p1, p2, p3 \in \big\{1, 2, \dots, N \big\} \) and \(N = p1 \neq p2 \neq p3\).
    
    \item \textbf{Combination operation:} Recombination (cross over) is applied after mutation operation. 
    \item \textbf{Selection operation:} This operation determines the offspring in the next generation should become a member of the population or not. 
    \item \textbf{Stopping criterion:} The current generation is substituted by the new generation until the criterion of termination is satisfied.
\end{enumerate}

\subsection{Harmony Search algorithm}
Harmony Search algorithm is proposed by Geem et al. in 2001 \cite{Lee}. This algorithm is inspired by the musical process of searching for a perfect state of harmony. Like harmony in music, solution vector of optimization and improvisation from the musician are analogous to structures of local and global search in optimization techniques.

In improvisation of the music, the players sound any pitch in the possible range together that can create one vector of harmony. In the case of pitches create a real harmony; this experience is stored in the memory of each player and they have the opportunity to create better harmony next time \cite{Lee}. There are three possible alternatives when one pitch is improvised by a musician: any one pitch is played from her/his memory, a nearby pitch is played from her/his memory and an entirely random pitch are played with the range of possible sound. If these options are used for optimization, they have three equivalent components; the use of harmony memory, pitch adjusting, and randomization. In HS algorithm, these rules are correlated with two relevant parameters, i.e. harmony consideration rate (HMCR) and pitch adjusting rate (PAR). The procedure of HS algorithm can be summarized into five steps as follows \cite{Lee}:

\begin{enumerate}
    \item \textbf{Initialize the problem and parameters:} In this algorithm, the problem can be maximum or minimum optimization, and the relevant parameters are HMCR, PAR, size of harmony memory and termination criterion.
    \item \textbf{Initialize harmony memory:} The harmony memory (HM) is usually initialized as a matrix that is created randomly as a vector of solution and arrange based on the objective function.
    \item \textbf{Improve a new harmony:} A vector of new harmony is produced from HM based on HMCR, PAR, and randomization. Selection of new value based on HMCR parameter by range 0 and 1. The vector of new harmony is observed to decide whether it should be pitch-adjusted using PAR parameter. The process of pitch adjusting is executed only after a value is selected from HM. 
    \item \textbf{Update harmony memory:} The new harmony substitutes the worst harmony in terms of the value of the fitness function, in which the fitness function of new harmony is better than worst harmony. 
    \item \textbf{Repeat (3) and (4) until satisfying the termination criterion:} In the case of meeting the termination criterion, the computation is ended. Alternatively, process (3) and (4) are reiterated. In the end, the vector of the best HM is nominated and is reflected as the best solution for the problem. 
\end{enumerate}

\section{Convolution Neural Network}
Convolution Neural Network is a variant of the standard multilayer perceptron (MLP). A substantial advantage of this method, especially for pattern recognition compared with conventional approaches is due to its capability in reducing the dimension of data, extracting the feature sequentially, and classifying in one structure of network \cite{Bengio}. The basic architecture model of CNN is inspired in 1962, from visual cortex proposed by Hubel and Wiesel. 

In 1980, Fukushima’s Neocognitron created the first computation of this model, and then in 1989, following the idea of Fukushima, LeCun et al. found the state-of-the-art performance on a number of tasks for pattern recognition using error gradient method \cite{LeCun}. 

The classical CNN by LeCun et al. is an extension of traditional MLP based on three ideas: local receive fields, weights sharing, and spatial/temporal sub-sampling. These ideas can be organized into two types of layers, which are convolution layers and subsampling layers. As is showed in Fig.1, the processing layers contain three convolution layers C1, C3, and C5, combined in between with two sub-sampling layers S2 and S4, and output layer F6. These convolution and sub-sampling layers are structured into planes called features maps.

In convolution layer, each neuron is linked locally to a small input region (local receptive field) in the preceding layer. All neurons with similar feature maps obtain data from different input regions until the whole of plane input is skimmed, but the same of weights is shared (weights sharing).

\begin{figure}
\includegraphics[scale = 0.6]{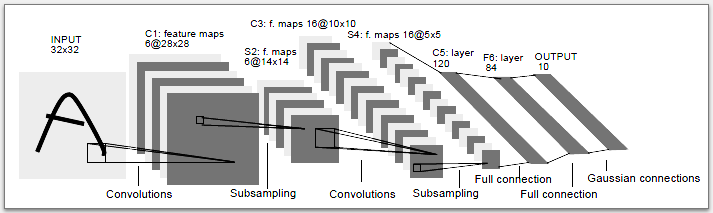}
\caption{Architecture of CNN by LeCun et al (LeNet-5)}
\label{fig:my_label}
\end{figure}

In sub-sampling layer, the feature maps are spatially down-sampled, in which the size of the map is reduced by a factor 2. As an example, the feature map in layer C3 of size 10x10 is sub-sampled to a conforming feature map of size 5x5 in the subsequent layer S4. The last layer is F6 that is the process of classification \cite{LeCun}.

Principally, a convolution layer is correlated with some feature maps, the size of the kernel, and connections to the previous layer. Each feature maps is the results of a sum of convolution from the maps of the previous layer, by their corresponding kernel and a linear filter. Adding a bias term and applying it to a non-linear function. The k-th feature map \(M_{ij}^k\) with the weights \(W^k\) and bias \(b_k\) is obtained using the $\tanh$ function as follow:

\begin{equation}
M_{ij}^k=\tanh((W^k \times x)_{ij} + b_k)
\end{equation}

The purpose of a sub-sampling layer is to reach spatial invariance by reducing the resolution of feature maps, in which each pooled feature map relates to one feature map of the preceding layer. The sub-sampling function, where \(a_i^{n \times n} \) is the inputs, \( \beta \) is a trainable scalar, and \( b \) is trainable bias, is given by the following equation:

\begin{equation}
a_j=\tanh\left(\beta\sum_{N\times N}{a_i^{n \times n} + b}\right)
\end{equation}

After several convolution and sub-sampling, the last structure is classification layer. This layer works as an input for a series of fully connected layers that will execute the classification task. It has one output neuron every class label, and in the case of MNIST dataset, this layer contains ten neurons corresponds to their classes.

\section{Design of proposed methods}
The architecture of this proposed method refers to a simple CNN structure (LeNet-5), not a complex structure like AlexNet\cite{Alexnet}. We use two variations of design structure. First is i-6c-2s-12c-2s, where the number of C1 is 6, and C2 is 12. Second is i-8c-2s-16c-2s, where the number of C1 is 8 and C2 is 18. The kernel size of all convolution layer is 5x5,  and the scale of sub-sampling is 2.These architecture is designed for recognizing handwritten digits from MNIST dataset.

In this proposed method, SA, DE and HS algorithm are used to train CNN (CNNSA, CNNDE, CNNHS) to find the condition of best accuracy and also to minimize estimated error and indicator of network complexity. This objective can be realized by computing the lost function of vector solution or the standard error on the training set. The following is the lost function used in this paper:
\begin{equation}
y= \frac {1}{2} \left({\frac{\sum_{i=N}^{N}{(o - u)^2}}{N}}\right)^{0.5}
\end{equation}
where $o$ is the expected output, $u$ is the real output and $N$ is some training samples. In the case of termination criterion, two situations are used in this method. The first is when the maximum iteration has been reached and the second is when the loss function is less than a certain constant. Both conditions mean that the most optimal state has been achieved.

\subsection{Design of CNNSA method}

Principally, algorithm on CNN computes the values of weight and bias,  in which on the last layer they are used to calculate the lost function. These values of weight and bias in the last layer are used as solution vector, denoted as $x$, to be optimized in SA algorithm, by adding $\Delta x$ randomly. 

The $\Delta x$ is the essential aspect of this proposed method. Selection in the proper of this value will significantly increase the accuracy. For example in CNNSA to one epoch, if $\Delta x = 0.0008 \times$ rand, then the accuracy is 88.12, in which this value is 5.73 greater than the original CNN (82.39). However, if $\Delta x = 0.0001 \times$ rand, its accuracy is 85.79 and its value is only 3.40 greater than the original CNN.

Furthermore, this solution vector is updated based on SA algorithm. When the termination criterion is satisfied, all of weights and biases are updated for all layers in the system. The following is the CNNSA algorithm of the proposed method.

\begin{algorithm}[H]
\SetAlgoLined
\KwResult{accuracy, time}
 initialization and set-up: i-6c-2s-12c-2s \;
 calculation process: weights ($w$), biases ($b$), lost function \(f(x)\)\; 
 solution vector ($x$): $w$ and $b$ on the last layer\; 
 
 \While{termination criteria is not satisfied}{
 \For{number of x'}{
  \( x' = x + \Delta x, f(x')\)\;
  \eIf{\( f(x') \leq f(x)\)}{
   \( x  \leftarrow  x'\)\;
   }{
   \( x \leftarrow x'\) with a transition probability ($p$)\;
  }
  }
  decrease the temperature: \(T=c \times T\) \;
  update $x$  for all layer;
 }
 
 \caption{CNNSA}
\end{algorithm}

\subsection{Design of CNNDE method}

At the first time, this method computes all the values of weight and bias. The values of weight and bias on the last layer ($x$) are used to calculate the lost function, and then by adding $\Delta x$ randomly, these new values are used to initialize the individuals in the population. 

\vspace{3mm}

\begin{algorithm}[H]
\SetAlgoLined
\KwResult{accuracy, time}
 initialization and set-up: i-6c-2s-12c-2s \;
 calculation process: weights ($w$), biases ($b$), lost function \(f(x)\)\;
 individual \(x_M^i\) in population: $w$ and $b$ on the last layer\; 
 
\While{termination criteria is not satisfied}{
\For{each of individual \(x_M^i\) in population \(P_M\)}{
   select auxiliary parents \( x_M^1, x_M^2, x_M^3\)\;
   create offspring \(x_M^{child}\) using mutation and recombination\;
   \(P_{M+1}=P_{M+1} \cup \) Best \((x_M^{child},x_M^i)\)\;    
}
M=M + 1 \;
update $x$ for all layer;
}

\caption{CNNDE}
\end{algorithm}

\vspace{3mm}
Similar to CNNSA method, selection in the proper of $\Delta x$ will significantly increase the value of accuracy. In the case of one epoch in CNNDE as an example, if $\Delta x = 0.0008 \times$ rand, then the accuracy is 86.30, in which this value is 3.91 greater than the original CNN (82.39). However, if $\Delta x = 0.00001 \times$ rand, its accuracy is 85.51.

Furthermore, these individual in the population are updated based on the of DE algorithm. When the termination criterion is satisfied, all of weights and biases are updated for all layers in the system. The following is the CNNDE algorithm of the proposed method.

\subsection{Design of CNNHS method}

At the first time like CNNSA and CNNDE, this method computes all the values of weight and bias. The values of weight and bias on the last layer ($x$) are used to calculate the lost function, and then by adding $\Delta x$ randomly, these new values are used to initialize the harmony memory. 

In this method, $\Delta x$ is also an important aspect, while selection the proper of $\Delta x$ will significantly increase the value of accuracy. For example of one epoch in CNNHS (i-8c-2s-16c-2s), if $\Delta x = 0.0008 \times$ rand, then the accuracy is 87.23, in which this value is 7.14 greater than the original CNN (80.09). However, if $\Delta x = 0.00001 \times$ rand, its accuracy is 80.23, the value is only 0.14 greater than CNN.

Furthermore, these harmony memory is updated based on the HS algorithm. When the termination criterion is satisfied, all of weights and biases are updated for all layers in the system. The following is the CNNHS algorithm of the proposed method.

\vspace{3mm}

\begin{algorithm}[H]
\SetAlgoLined
\KwResult{accuracy, time}
 initialization and set-up: i-6c-2s-12c-2s \;
 calculation process: weights ($w$), biases ($b$), lost function \(f(x)\)\; 
 harmony memory \(x_M^i \):  $w$ and $b$ on the last layer\; 
\While{termination criterion is not satisfied}{
\For{number of search}{
\eIf{\(rand < HMCR\)}{
    \(x_{new}^i \)from HM  \;
  }{
  \eIf{\(rand<PAR\) }{
    \(x_{new}^i = x_{new}^i + \Delta x\)}
  
  }
}
    \(x_{new}^i=x_{min} + rand(x_{max} - x_{min}) \)

}
\caption{CNNHS}
\end{algorithm}

\section{Simulation and results}
In this paper, the primary goal is to improve the accuracy of original CNN by using SA, DE, and HS algorithm. This can be performed by minimizing the classification task error tested on the MNIST dataset. Some of the examples image for MNIST dataset are shown in Fig.2.

\begin{figure}
\includegraphics[scale=0.8]{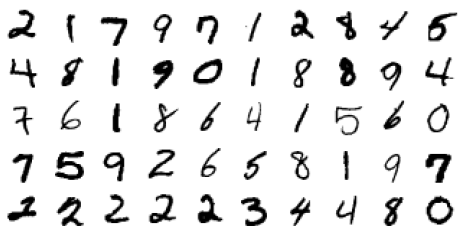}
\caption{Example of some images from MNIST data-set}
\label{fig:my_label}
\end{figure}

In CNNSA experiment, the size of neighborhood was set = 10 and maximum of iteration (maxit) = 10. In CNNDE, the population size = 10 and  maxit = 10. In CNNHS, the harmony memory size = 10 and maxit = 10. Since it is difficult to make sure the control of parameter, in all of the experiment the values of c = 0.5 for SA, F = 0.8 and cr = 0.3 for DE, as well as HMCR = 0.8 and PAR = 0.3 for HS. We also set the parameter of CNN, i.e., the learning rate ($\alpha = 1$) and the batch size (100). 

As for the epoch parameter, the number of epoch 1 to 10 for every experiment. All of the experiment was implemented in MATLAB-R2011a, on a personal computer with processor Intel Core i7-4500u, 8 GB RAM running memory, in Window 10, with five separate runtimes. The original program of this simulation is DeepLearn Toolbox from Palm\cite{IMM2012-06284}.

\begin{table}[]
\small
\caption{Accuracy (Acc.) and its standard deviation (Std.Dev) for design: i-2s-6c-2s-12c}
\label{tab:my_label}

\begin{tabular}{c c c c c c c c c}
\hline
\multicolumn {1}{c}{\multirow{2}{*}{Epoch}} & \multicolumn{2}{c}{CNN} &
\multicolumn {2}{c}{CNNSA} & \multicolumn {2}{c}{CNNDE} & \multicolumn {2}{c}{CNNHS}\\
\cline {2-9}

\multicolumn {1}{r}{}       & \multicolumn {1}{c}{Acc.} & \multicolumn {1}{c}{Std.Dev.} &
\multicolumn {1}{c}{Acc.} & \multicolumn {1}{c}{Std.Dev.}  & \multicolumn {1}{c}{Acc.} &
\multicolumn {1}{c}{Std.Dev.} & \multicolumn {1}{c}{Acc.} & \multicolumn {1}{c}{Std.Dev.} \\
\hline

1		& 82.39 & n/a	 & 88.12 & 0.39	  & 86.30 & 0.33    & 87.23  & 0.95  \\
2 		& 89.06 & n/a	 & 92.77 & 0.43	  & 91.33 & 0.19    & 91.20  & 0.33  \\
3		& 91.13 & n/a	 & 94.61 & 0.31	  & 93.45 & 0.28  	& 93.24  & 0.40  \\
4		& 92.33 & n/a	 & 95.57 & 0.16	  & 94.63 & 0.44 	& 93.77  & 0.12  \\
5		& 93.11	& n/a    & 96.29 & 0.14	  & 95.15 & 0.15    & 94.89  & 0.33 \\
6		& 93.67 & n/a	 & 96.61 & 0.18   & 95.67 & 0.20  	& 95.17  & 0.43 \\
7 		& 94.25	& n/a    & 96.72 & 0.12   & 96.28 & 0.20    & 95.65  & 0.20 \\
8		& 94.77 & n/a	 & 96.99 & 0.11	  & 96.59 & 0.11 	& 96.08  & 0.24 \\
9		& 95.37 & n/a	 & 97.11 & 0.06   & 96.68 & 0.17 	& 96.16  & 0.11 \\
10		& 95.45 & n/a	 & 97.37 & 0.14	  & 96.86 & 0.10	& 96.98  & 0.06 \\

\hline
\end{tabular}
\end{table}

\begin{table}[]
\small
\caption{Computation time and its Std.Dev. for design: i-2s-6c-2s-12c}
\label{tab:my_label}

\begin{tabular}{c c c c c c c c c}
\hline
\multicolumn {1}{c}{\multirow{2}{*}{Epoch}} & \multicolumn{2}{c}{CNN} &
\multicolumn {2}{c}{CNNSA} & \multicolumn {2}{c}{CNNDE} & \multicolumn {2}{c}{CNNHS}\\
\cline {2-9}

\multicolumn {1}{r}{}       & \multicolumn {1}{c}{Time} & \multicolumn {1}{c}{Std.Dev.} &
\multicolumn {1}{c}{Time} & \multicolumn {1}{c}{Std.Dev.}  & \multicolumn {1}{c}{Time} &
\multicolumn {1}{c}{Std.Dev.} & \multicolumn {1}{c}{Time} & \multicolumn {1}{c}{Std.Dev.} \\
\hline

1		& 93.21  & n/a    & 117.48  & 1.12  & 138.58  & 0.90    & 160.92  & 0.85  \\
2 		& 225.05 & n/a    & 243.43  & 9.90  & 278.08  & 1.66    & 370.59  & 5.87  \\
3		& 318.84 & n/a    & 356.49  & 1.96  & 414.64  & 2.43    & 414.13  & 0.63  \\
4		& 379.44 & n/a    & 479.83  & 1.95  & 551.39  & 2.28 	& 554.51  & 0.73  \\
5		& 479.04 & n/a    & 596.35  & 4.08  & 533.21  & 1.42    & 692.90  & 2.90  \\
6		& 576.38 & n/a    & 721.48  & 1.48  & 640.30  & 6.19    & 829.56  & 1.95  \\
7 		& 676.57 & n/a    & 839.55  & 1.19  & 744.89  & 3.98    & 968.18  & 1.97  \\
8		& 768.24 & n/a    & 960.69  & 1.74  & 852.74  & 4.48    & 1105.2  & 1.39  \\
9		& 855.85 & n/a    & 1082.18 & 2.54  & 957.89  & 5.78    & 1245.54 & 4.96  \\
10		& 954.54 & n/a    & 1202.52 & 2.08  & 1373.1  & 1.51    & 1623.13 & 4.36 \\

\hline
\end{tabular}
\end{table}

\begin{figure}
\includegraphics[scale=0.4]{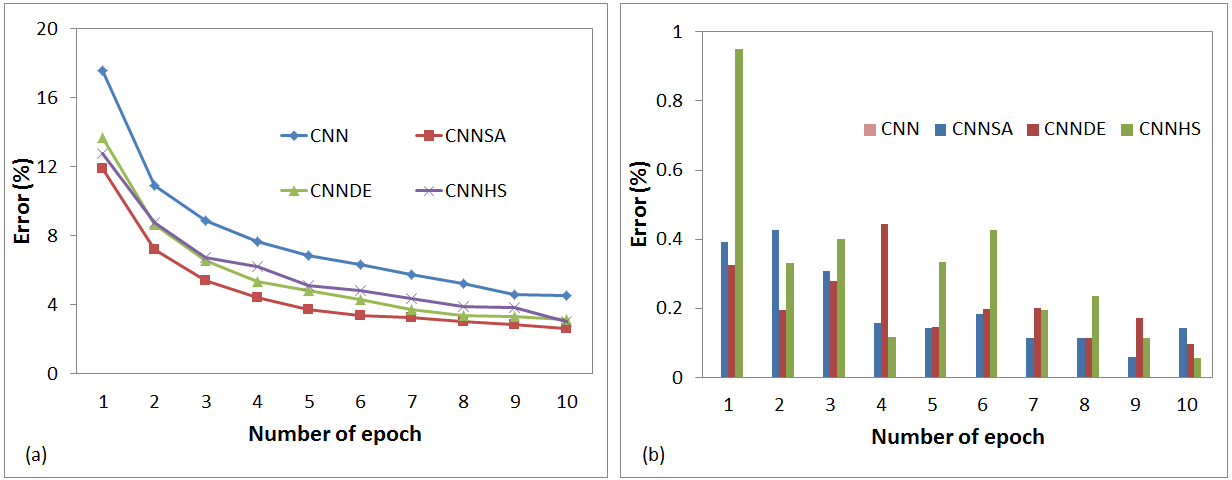}
\caption{Error and its Std.Dev. (i-6c-2s-12c-2s)}
\label{fig:my_label}
\end{figure}

\begin{figure}
    \includegraphics[scale=0.4]{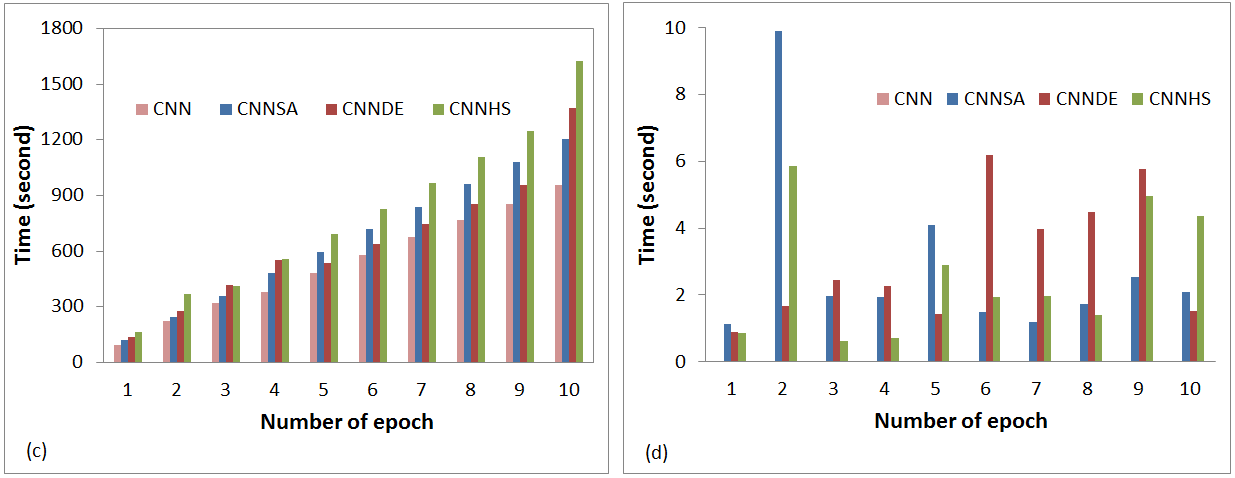}
    \caption{Computation time and its Std.Dev. (i-6c-2s-12c-2s)}
    \label{fig:my_label}
\end{figure}

\begin{table}[]
\small
\caption{Accuracy and its Std.Dev. for design: i-2s-8c-2s-16c}
\label{tab:my_label}

\begin{tabular}{c c c c c c c c c}
\hline
\multicolumn {1}{c}{\multirow{2}{*}{Epoch}} & \multicolumn{2}{c}{CNN} &
\multicolumn {2}{c}{CNNSA} & \multicolumn {2}{c}{CNNDE} & \multicolumn {2}{c}{CNNHS}\\
\cline {2-9}

\multicolumn {1}{r}{}       & \multicolumn {1}{c}{Acc.} & \multicolumn {1}{c}{Std.Dev.} &
\multicolumn {1}{c}{Acc.} & \multicolumn {1}{c}{Std.Dev.}  & \multicolumn {1}{c}{Acc.} &
\multicolumn {1}{c}{Std.Dev.} & \multicolumn {1}{c}{Acc.} & \multicolumn {1}{c}{Std.Dev.} \\
\hline

1		& 80.09 & n/a	 & 86.36 & 0.76	  & 84.78 & 1.24    & 87.23  & 0.57  \\
2 		& 89.04 & n/a	 & 91.18 & 0.25	  & 91.63 & 0.30    & 92.15  & 0.55  \\
3		& 90.98 & n/a	 & 93.56 & 0.20	  & 93.67 & 0.17  	& 93.69  & 0.31  \\
4		& 92.27 & n/a	 & 94.69 & 0.16	  & 94.86 & 0.43 	& 94.63  & 0.20  \\
5		& 93.17	& n/a    & 95.51 & 0.12	  & 95.57 & 0.04    & 95.30  & 0.16 \\
6		& 93.79 & n/a	 & 96.23 & 0.08   & 96.20 & 0.14  	& 95.80  & 0.25 \\
7 		& 94.74	& n/a    & 96.52 & 0.08   & 96.52 & 0.32    & 95.71  & 0.24 \\
8		& 95.22 & n/a	 & 96.95 & 0.07	  & 96.68 & 0.19 	& 96.40  & 0.13 \\
9		& 95.54 & n/a	 & 97.18 & 0.08   & 97.10 & 0.00 	& 96.84  & 0.27 \\
10		& 96.05 & n/a	 & 97.35 & 0.02	  & 97.32 & 0.04	& 96.77  & 0.04 \\

\hline
\end{tabular}
\end{table}

\begin{figure}
    \includegraphics[scale=0.4]{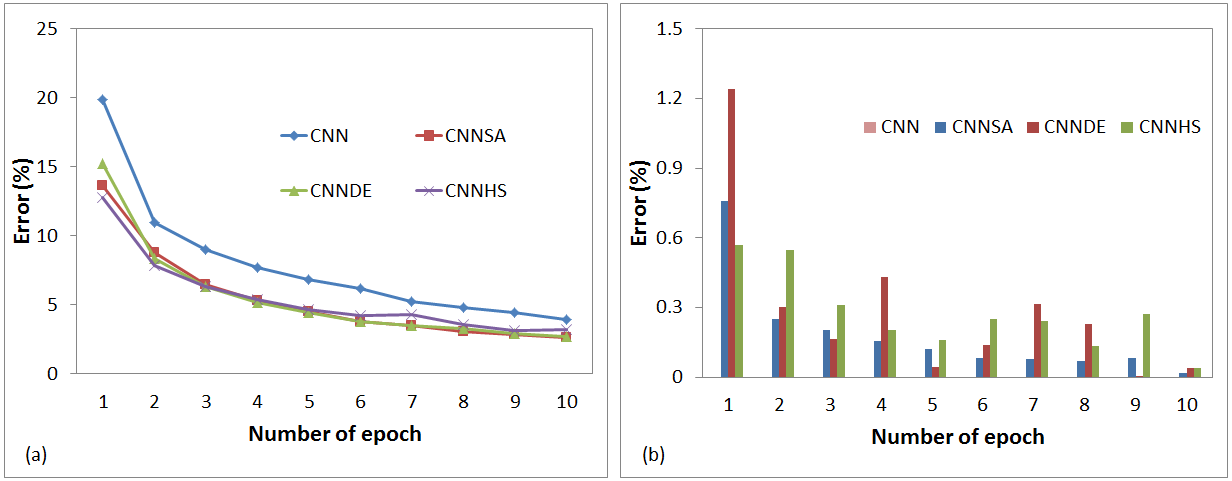}
    \caption{Error and its Std.Dev. (i-8c-2s-16c-2s) }
    \label{fig:my_label}
\end{figure}

\begin{figure}
\includegraphics[scale=0.4]{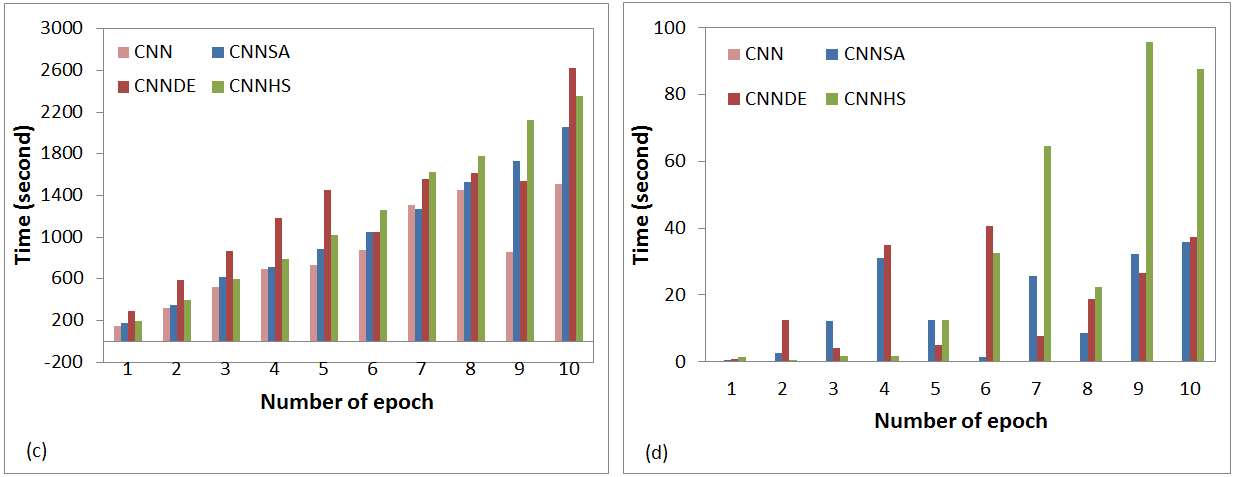}
\caption{Computation time and its Std.Dev. (i-8c-2s-16c-2s)}
\label{fig:my_label}
\end{figure}

\begin{table}[]
\small
\caption{Computation time and its Std.Dev. for design: i-2s-8c-2s-16c}
\label{tab:my_label}

\begin{tabular}{c c c c c c c c c}
\hline
\multicolumn {1}{c}{\multirow{2}{*}{Epoch}} & \multicolumn{2}{c}{CNN} &
\multicolumn {2}{c}{CNNSA} & \multicolumn {2}{c}{CNNDE} & \multicolumn {2}{c}{CNNHS}\\
\cline {2-9}

\multicolumn {1}{r}{}       & \multicolumn {1}{c}{Time} & \multicolumn {1}{c}{Std.Dev.} &
\multicolumn {1}{c}{Time} & \multicolumn {1}{c}{Std.Dev.}  & \multicolumn {1}{c}{Time} &
\multicolumn {1}{c}{Std.Dev.} & \multicolumn {1}{c}{Time} & \multicolumn {1}{c}{Std.Dev.} \\
\hline

1		& 145.02  & n/a    & 175.08  & 0.64  & 289.54  & 0.78    & 196.10  & 1.35  \\
2 		& 323.62  & n/a    & 353.55  & 2.69  & 586.96  & 12.56   & 395.43  & 0.60  \\
3		& 520.16  & n/a    & 614.71  & 12.10 & 868.82  & 4.02    & 597.391 & 1.83  \\
4		& 692.80  & n/a    & 718.53  & 31.05 & 1185.49 & 34.95   & 794.43  & 1.70  \\
5		& 729.05  & n/a    & 885.64  & 12.53 & 1451.64 & 4.99    & 1023.72 & 12.51 \\
6		& 879.17  & n/a    & 1051.30 & 1.30  & 1045.26 & 40.62   & 1255.93 & 32.54 \\
7 		& 1308.21 & n/a    & 1271.03 & 25.55 & 1554.67 & 7.86    & 1627.30 & 64.56  \\
8		& 1455.06 & n/a    & 1533.30 & 8.55  & 15.39   & 106.75  & 1773.92 & 2251  \\
9		& 1392.62 & n/a    & 1726.50 & 32.31 & 1573.52 & 18.42   & 2123.32 & 95.76  \\
10		& 1511.74 & n/a    & 2054.40 & 35.85 & 2619.62 & 37.37   & 2354.90 & 87.68 \\

\hline
\end{tabular}
\end{table}

\begin{figure}
    \includegraphics[scale=0.5]{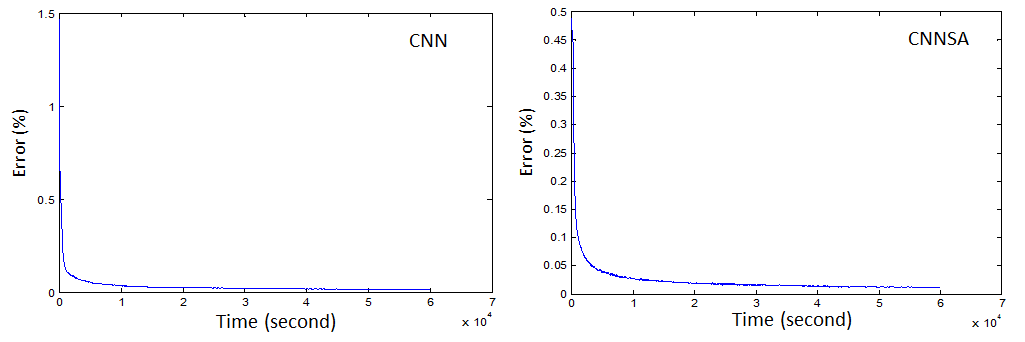}
    \includegraphics[scale=0.5]{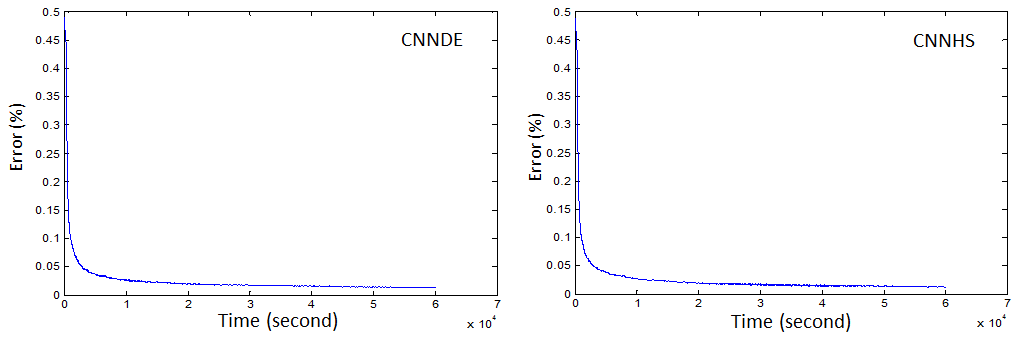}
    \caption{Error vs computation time for 100 epoch}
    \label{fig:my_label}
\end{figure}

All of the experiment results of the proposed methods are compared with the experiment result from the original CNN. These results for the design of i-6c-2s-12c-2s are summarized in Table 1 for accuracy, Table 2 for the computational time, Fig. 3 for error and its standard deviation as well as Fig. 4 for computational time and its standard deviation. The results for the design of i-8c-2s-16c-2s are summarized in Table 3 for accuracy, Table 4 for the computational time, Fig. 4 for error and its standard deviation as well as Fig. 5 for computational time and its standard deviation. 

The experiments of original CNN are conducted at only one time for each epoch because the value of its accuracy will not change if the experiment is repeated with the same condition. In general, the tests conducted showed that to the higher epoch value, the better is the accuracy. For example in one epoch, compared to CNN (82.39), the accuracy increased to 5.73 for CNNSA (88.12), 3.91 to CNNDE (86.30), and 4.84 to for CNNHS (87.23). While in 5 epoch, compared to CNN (93.11), the increase of accuracy is 3.18 for CNNSA (96.29), 2.04 for CNNDE (94.15), and 1.78 for CNNHS (94.89). In the case of 100 epoch, as shown in Fig.6, the increase in accuracy compared to CNN (98.65) is only 0.16 for CNNSA (98.81), 0.13 for CNNDE (98.78), and 0.09 for CNNHS (98.74). 

The experiment results show that CNNSA presents the best accuracy for all epoch. Accuracy improvement of CNNSA, compared to the original CNN, varies for each epoch, with a range of values between 1.74 (9 epoch) up to 5.73 (1 epoch). The computation time for the proposed method, compared to the original CNN, is in the range of 1.01 times (CNNSA, two epoch: 246/244) up to 1.70 times (CNNHS, nine epoch: 1246/856).

In addition, we also test our proposed method with CIFAR10 (canadian institute for advanced research)  data-set. This dataset consists of 60.000 color images, in which the size of every image is 32x32. There are five batches for training, composed of 50.000 images, and one batch of test images consist of 10.000 images. The CIFAR10 dataset is divided into ten classes, where each class has 6.000 images. Some example images of this dataset are showed in Fig.8 as follow.

\begin{figure}
    \includegraphics[scale=0.8]{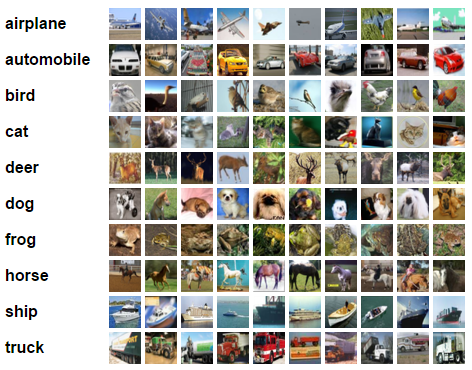}
    \caption{Example of some images from Cifar data-set}
    \label{fig:my_label}
\end{figure}

The experiment of CIFAR10 dataset was conducted in MATLAB-R2014a. We use the number of epoch 1 to 15 for this experiment. The original program is MatConvNet from \cite{vedaldi15matconvnet}. In this paper, the program was modified with SA algorithm. The results can be seen in Fig. 9 for objective, Fig.10 for top-1 error, and Fig.11 for a top-5 error. In general, these results show that CNNSA works better than original CNN for CIFAR10 data-set.

\begin{table}[]
\small
\caption{Comparison of CNN and CNNSA for train}
\label{tab:my_label}

\begin{tabular}{c c c c c c c}
\hline
\multicolumn {1}{c}{\multirow{2}{*}{Epoch}} & \multicolumn{3}{c}{CNN} &
\multicolumn {3}{c}{CNNSA}\\
\cline {2-7}

\multicolumn {1}{r}{}     & \multicolumn {1}{c}{Objective} & \multicolumn {1}{c}{Top-1 error} & \multicolumn {1}{c}{Top-5 error} &
\multicolumn {1}{c}{Objective} & \multicolumn {1}{c}{Top-1 error} & \multicolumn {1}{c}{Top-5 error}\\
\hline

1		& 1.4835700   & 0.10676    & 0.52868    & 0.009493  & 0.00092    & 0.00168        \\
2 		& 1.0443820   & 0.03664    & 0.36148    & 0.013218  & 0.00094    & 0.00188         \\
3		& 0.9158232   & 0.02686    & 0.31518    & 0.010585  & 0.00094    & 0.0017        \\
4		& 0.8279042   & 0.02176    & 0.28358    & 0.008023  & 0.00096    & 0.00188   	  \\
5		& 0.7749367   & 0.01966    & 0.26404    & 0.009285  & 0.00106    & 0.00186        \\
6		& 0.7314783   & 0.01750    & 0.25076    & 0.013674  & 0.00102    & 0.00175        \\
7 		& 0.6968027   & 0.01566    & 0.23968    & 0.117740  & 0.0009     & 0.00168        \\
8		& 0.6654411   & 0.01398    & 0.22774    & 0.011239  & 0.0011     & 0.0018        \\
9		& 0.6440073   & 0.01320    & 0.21978    & 0.011338  & 0.00106    & 0.0018        \\
10		& 0.6213060   & 0.01312    & 0.20990    & 0.009957  & 0.00116    & 0.0019        \\
11		& 0.6024042   & 0.01184    & 0.20716    & 0.008434  & 0.00096    & 0.00176        \\
12 		& 0.5786811   & 0.01090    & 0.19954    & 0.009425  & 0.0011     & 0.00192          \\
13		& 0.5684009   & 0.01068    & 0.19548    & 0.012485  & 0.00082    & 0.0018        \\
14		& 0.5486258   & 0.00994    & 0.18914    & 0.012108  & 0.00098    & 0.00184   	  \\
15		& 0.5347288   & 0.00986    & 0.18446    & 0.009675  & 0.0011     & 0.00186       \\

\hline
\end{tabular}
\end{table}

\begin{table}[]
\small
\caption{Comparison of CNN and CNNSA for validation}
\label{tab:my_label}

\begin{tabular}{c c c c c c c}
\hline
\multicolumn {1}{c}{\multirow{2}{*}{Epoch}} & \multicolumn{3}{c}{CNN} &
\multicolumn {3}{c}{CNNSA}\\
\cline {2-7}

\multicolumn {1}{r}{}     & \multicolumn {1}{c}{Objective} & \multicolumn {1}{c}{Top-1 error} & \multicolumn {1}{c}{Top-5 error} &
\multicolumn {1}{c}{Objective} & \multicolumn {1}{c}{Top-1 error} & \multicolumn {1}{c}{Top-5 error}\\
\hline

1		& 1.148227   & 0.0466    & 0.3959    & 0.034091  & 0.0039    & 0.0087        \\
2 		& 0.985902   & 0.0300    & 0.3422    & 0.061806  & 0.0044    & 0.0091         \\
3		& 0.873938   & 0.0255    & 0.2997    & 0.054007  & 0.0050    & 0.0091        \\
4		& 0.908667   & 0.0273    & 0.3053    & 0.054711  & 0.0051    & 0.0091   	  \\
5		& 0.799778   & 0.0226    & 0.2669    & 0.043632  & 0.0044    & 0.0091        \\
6		& 0.772151   & 0.0209    & 0.2614    & 0.071143  & 0.0057    & 0.0091        \\
7 		& 0.784206   & 0.0210    & 0.2593    & 0.065040  & 0.0050    & 0.0095        \\
8		& 0.732094   & 0.0170    & 0.2474    & 0.048466  & 0.0061    & 0.0095        \\
9		& 0.761574   & 0.0217    & 0.2532    & 0.056708  & 0.0056    & 0.0091        \\
10		& 0.763323   & 0.0207    & 0.2515    & 0.044423  & 0.0048    & 0.0086        \\
11		& 0.720129   & 0.0165    & 0.2352    & 0.047963  & 0.0041    & 0.0087        \\
12 		& 0.700847   & 0.0167    & 0.2338    & 0.063033  & 0.0055    & 0.0087          \\
13		& 0.729708   & 0.0194    & 0.2389    & 0.068989  & 0.0052    & 0.0096        \\
14		& 0.747789   & 0.0192    & 0.2431    & 0.056425  & 0.0049    & 0.0091   	  \\
15		& 0.723088   & 0.0182    & 0.2355    & 0.052753  & 0.0052    & 0.0095       \\

\hline
\end{tabular}
\end{table}

\begin{figure}
    \includegraphics[scale=0.65]{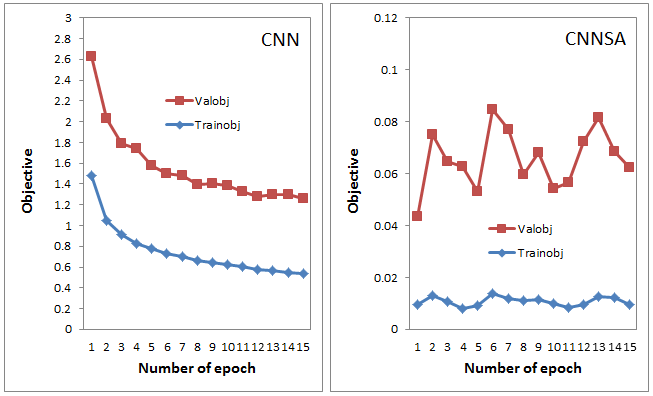}
    \caption{CNN vs CNNSA for Objective}
    \label{fig:my_label}
\end{figure}

\begin{figure}
    \includegraphics[scale=0.65]{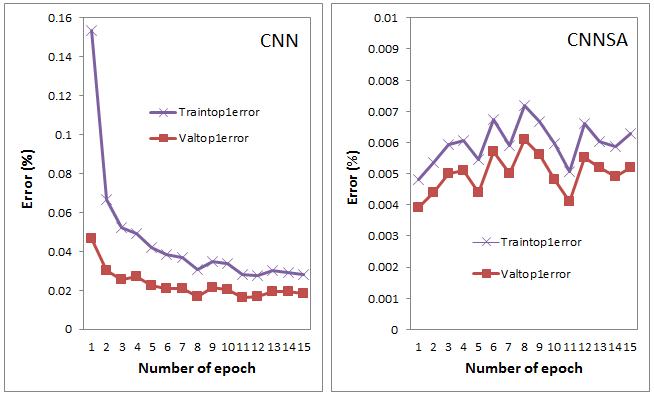}
    \caption{CNN vs CNNSA for Top-1 error}
    \label{fig:my_label}
\end{figure}

\begin{figure}
    \includegraphics[scale=0.65]{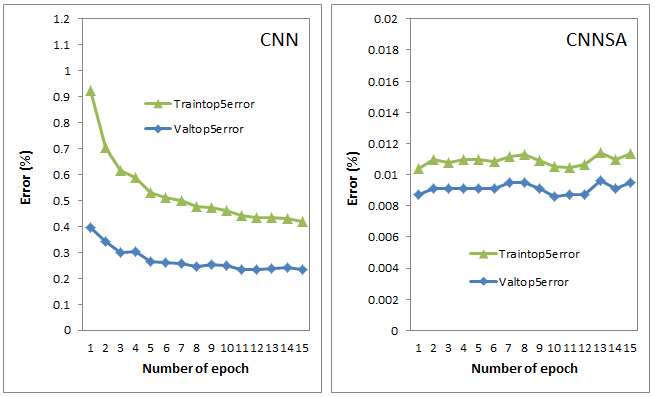}
    \caption{CNN vs CNNSA for Top-5 error}
    \label{fig:my_label}
\end{figure}

\section{Conclusion}
This paper shows that SA, DE and HS algorithms improve the accuracy of the CNN. Although there is an increase in computation time, nevertheless error of the proposed method is smaller than the original CNN for all variation of the epoch.

It is possible to validate the performance of this proposed method on other benchmark datasets such as ORL, INRIA, Hollywood II, and ImageNet. This strategy can also be developed for other metaheuristic algorithms such as ACO, PSO, and BCO to optimize CNN. 

For the future study, metaheuristic algorithms applied to the other DL methods need to be explored, such as the recurrent neural network, deep belief network, and AlexNet (a newer variant of CNN).


\bibliography{library}

\bibliographystyle{abbrv}

\end{document}